*Review*

# A Critical Analysis of Patch Similarity Based Image Denoising Algorithms

Varuna De Silva*
Loughborough University London

Correspondence: V.D.De-Silva [at] lboro.ac.uk

**Abstract:** Image denoising is a classical signal processing problem that has received significant interest within the image processing community during the past two decades. Most of the algorithms for image denoising has focused on the paradigm of non-local similarity, where image blocks in the neighborhood that are similar, are collected to build a basis for reconstruction. Through rigorous experimentation, this paper reviews multiple aspects of image denoising algorithm development based on non-local similarity. Firstly, the concept of "non-local similarity" as a foundational quality that exists in natural images has not received adequate attention. Secondly, the image denoising algorithms that are developed are a combination of multiple building blocks, making comparison among them a tedious task. Finally, most of the work surrounding image denoising presents performance results based on Peak-Signal-to-Noise Ratio (PSNR) between a denoised image and a reference image (which is perturbed with Additive White Gaussian Noise). This paper starts with a statistical analysis on "non-local similarity" and its effectiveness under various noise levels, followed by a theoretical comparison of different state-of-the-art image denoising algorithms. Finally, we argue for a methodological overhaul to incorporate no-reference image quality measures and unprocessed images (raw) during performance evaluation of image denoising algorithms.

## 1. Introduction

With the emergence of the widespread use of smartphones, capturing images and videos has become ubiquitous in our daily lives. Advancement of sensor technology means now we are capable of capturing high resolution images at a very high quality. The image signal processing pipeline of an imaging device plays a crucial role in delivering a high quality image/video frame through processing and cleaning the raw sensor data that is captured by the sensor utilizing image processing algorithms. Demosaicking, noise reduction, shading correction, white balance correction, gamma correction and dynamic range compression are key components of a state-of-the-art Image Signal Processor (ISP). These functions of an ISP deals with compensating for limitations of the sensing technology.

The objective of image denoising algorithms is to reconstruct the original scene that was captured by the device. In doing so, the algorithm has to detect and mitigate the effects of imperfections in the sensing process. The image denoising algorithms can be broadly categorized in to two classes: spatial domain methods that exploit the pixel correlations that exist in natural images and transform domain methods that exploit the correlations of transformed coefficients in a sparse domain. The spatial domain methods include Gaussian filtering, bilateral filtering [1], anisotropic filtering [2] or steering kernel regression based filtering [3]. In 2005, a new spatial domain denoising technique named as Non-Local Means denoising (NLM) was proposed by Buades et al.[4], which estimates each pixel by weighted averaging of all pixels in the image. The weights in [4] were derived based on the Euclidean distance between the patch centred around the pixel being denoised and the patch centred around a neighbourhood pixel. The transform domain methods on the other hand represent image partitions as a function of an orthonormal basis such as wavelets [5] or curvelets [6], and denoising is obtained by thresholding or shrinkage of the coefficients. The main assumption of transform domain methods is that while the noise energy is uniformly distributed across all the coefficients, the image content is concentrated in a few largest coefficients [7]. Different variations of such transform domain denoising techniques exist in literature such as, BayesShrink [8], ProbShrink [9], or Bivariate Shrinkage [10].

Since the introduction of NLM filter by Buades et al., many improvements and variants on this filter was proposed [11], [12], [13] [14], which exploited the structural redundancy that is observable in natural images. One such extension of the NLM filter, was the Block Matching 3D with collaborative filtering method (BM3D) [15], which to date is considered as the benchmark for comparing image denoising algorithms. Conceptually it can be thought that the in [15] NLM concept was combined with transform domain methods. According to [15], to denoise a given pixel, a patch surrounding the pixel is considered along with similar patches from the neighbourhood and transformed using a basis function, to perform coefficient shrinkage. Since the publication of [15], several image denoising frameworks were proposed [16][17] [18] [7], all of which exploited the structural redundancy in images and largely followed the same algorithmic steps of BM3D. While this branch of denoising algorithms have consistently outperformed traditional spatial filters and transform domain methods, they do not consistently outperform BM3D. More recently, Trainable Nonlinear Reactive Diffusion (TNRD) [19] method was proposed and a multitude of deep convolutional neural networks have been explored with seemingly successful results beyond BM3D [20]–[25]. Somewhat contradicting to the claims in recent papers, Plotz et al., illustrates that when performance is analysed on images with real noise, various recent techniques that perform well on synthetic noise are clearly outperformed by BM3D, most notably the TNRD method [19]. Despite the interesting concepts often presented in papers, one would wonder the reason for the observed performance plateau.

The objective of this paper is to critically evaluate the branch of denoising algorithms that exploit patch based structural redundancy in natural images. We will call this branch of techniques as "patch similarity based denoising algorithms". Initially the paper summarizes the major concepts that underpin patch similarity based denoising algorithms and compare and contrast between these algorithms. The suitability of the concept of structural redundancy of natural images for denoising is not without its doubters [26]. To answer such critics, in this paper, we investigate the statistics of structural redundancy in natural images. Finally, the paper argues for a fresh look at the way image denoising algorithms are evaluated. While appreciating the numerous recent efforts with deep learning architectures proposed for image denoising [20]–[25], we do not engage in a full comparison of these techniques. However, for the sake completeness we have selected a representative technique Deep Residual CNN (DnCNN) [21] in our analysis.

The major contributions of this paper can be identified as follows:
- Provide a theoretical and experimental comparison of the patch similarity-based image denoising algorithms. The image denoising algorithms are construed within a specific framework, and the distinguishing factors of these algorithms are described.
- The concept of structural redundancy also known as the non-local similarity in natural images is experimentally investigated to quantify the effectiveness of patch based denoising algorithms.
- The experimental validation procedure of potential image denoising algorithms is critically evaluated to reflect the needs in practical / industrial settings, which advocates for no-reference image quality assessment.

*1.1. Comparison with similar papers*

One of the earliest reviews of image denoising was presented by Buades et al [27], which compared many algorithms at that time along with the NLM. A short account of different variations of non-local means algorithms were presented in [28]. Peyman Milanfar in his extensive review of image filtering argues that most algorithms proposed in the recent past have been closely correlated [29]. The review in [29], covers most aspects of image filtering, including historical developments in image denoising. In the current paper, we are focused on image denoising algorithms only and in particular on a class of algorithms that are based on non-local patch similarity.

A comparison of denoising techniques is provided in [30]. In [30] authors focus on local image filters as well as non-local filtering techniques for denoising. In comparison, the current work while focusing on non-local patch-based techniques, extends the coverage of denoising algorithms. A recent review focused only on patch based techniques can be found in [31], with a strong focus on BM3D algorithm, but overlooked many competing algorithms that were published since BM3D. The most recent of the reviews on the topic of image denoising is presented in, where authors compare between spatial domain techniques [32], transform domain techniques and hybrid methods. A review focused on denoising algorithms for biomedical images is found in [33]. A recent review of only the deep learning architectures for denoising can be found in [34].

The current paper is not a comprehensive review of Image denoising, but focused, critical investigation of image denoising algorithms that utilize the non-local similarity as a feature. This branch of image denoising has consistently outperformed many other techniques in the past. In addition to providing a concise theoretical analysis of this branch of algorithms, we are looking at two major aspects that have not received adequate attention. In the previous works, the patch jittering effect and rare-patch effect has been mentioned but a deep quantitative analysis was not performed. Similarly, non-local similarity effect was termed elusive. On the other hand, image-denoising papers almost exclusively focus on additive Gaussian noise, which is somewhat different from the reality. To address these issues, this paper provides novel insights into natural image statistics for non-local patch similarity and a framework for quality evaluation of denoising algorithms.

**Table 1.** Summary of algorithms compared in the paper.

| Algorithm / Reference | Year | Abbreviation |
|---|---|---|
| Non-Local Means Denoising [4] | 2005 | NLM |
| Denoising based on over complete dictionaries [35] | 2006 | k-SVD |
| Block Matching 3D collaborative filtering [15] | 2007 | BM3D |
| Learned Simultaneous Sparse Coding [36] | 2009 | LSSC |
| Clustering based denoising with Locally Learned Dictionaries [37] | 2009 | k-LLD |
| BM3D-Shape Adaptive Principal Component Analysis [16] | 2009 | BM3D-SAPCA |
| Principal Component Analysis with Local Pixel Grouping [17] | 2010 | LPG-PCA |
| Patch based Locally Optimal Wiener Filter [14] | 2012 | PLOW |
| Two Direction Non-Local model for image denoising [38] | 2013 | TDNL |
| Spatially Adaptive Iterative Singular-value Thresholding [26] | 2013 | SAIST |
| Non-locally Centralized Sparse Representation [18] | 2013 | NCSR |
| Adaptive Regularization of Non-Local Means [39] | 2014 | AR-NLM |
| Weighted Nuclear Norm Minimization | 2014 | WNNM |
| Efficient Low Rank Approximation of SVD [7] | 2016 | LRA-SVD |
| Residual Learning of Deep CNN for Image Denoising [21] | 2017 | DnCNN |

1.1.1. Organization of the paper

The rest of the paper is organized as follows: Section II introduces the algorithms reviewed in this paper along with an experimental evaluation of the denoising performance of them. In Section III, the concept of structural redundancy in natural images as relevant to image denoising is statistically analysed. The section IV presents an evaluation of performance measurement techniques for image denoising algorithms. The section V concludes the paper with recommendations for future work.

**2. Comparison of Image Denoising Algorithms**

The Table 1 contains different algorithms discussed in this paper, along with the abbreviation of the algorithms used. The algorithms are arranged in the chronological order of publication.

*2.1. Comparison of reconstruction quality of different denoising algorithms*

In this section we will compare the performance of a number of state-of-the-art image denoising algorithms. While there are a multitude of algorithms in literature, we focus on denoising algorithms based on the concept of non-local similarity. Following the tradition of algorithmic performance comparison methodology in literature, we add AWGN noise to an image and perform denoising. The performance of denoising is measured utilizing two full reference image quality metrics that are commonly utilized in literature: PSNR and SSIM. Later on, in this paper, we will argue against this method of performance comparison in section IV. The purpose of this experiment is to provide a summary of the performance of denoising algorithms, and to identify the best performing algorithms.

In this experiment, 10 different images are corrupted with AWGN at a noise standard deviation of 5. A total of 10 different image denoising algorithms are applied on the noisy images. The quality of the resultant image of the denoising algorithm is measure as the PSNR and the SSIM index. The results of the denoising process are summarized in Table 2.

Furthermore, we repeat this experiment at different noise levels and the results are averaged across all the images. The denoising performance is presented in Figure 1 as the PSNR/SSIM against the noise variance. For clarity of presentation, we have selected a subset of the methods in table 1/2 for illustration in Figure 1 (a) and (b).

The PSNR results indicate that there is very little difference between the top performing algorithms (Eg. BM3D, NCSR and LRA-SVD). In terms of SSIM results, the LRA-SVD algorithm performs marginally better than the rest. While falling short in PSNR measurements, EPLL algorithm performs very well in terms of SSIM. The NCSR algorithm, on the other hand performs well in terms of PSNR but falls behind the rest at high noise levels. Similarly, LPG-PCA algorithm performs well at lower noise levels but falls behind as the noise level increase. On the other hand WNNM algorithm performs consistently better than any other algorithm considered in this study, both in terms of PSNR, and in terms of SSIM, and the DnCNN algorithm consistently performs worst at all noise levels compared to rest of the algorithms shown in Figure 1. The results from these two experiments indicate WNNM and BM3D are the best performing algorithms of the ones considered in this paper.

**Table 2**. Performance comparison of denoising algorithms based on PSNR and SSIM.

| Metric | Image | BM3D | DnCNN | KLLD | PLOW | EPLL | WNNM | LPG-PCA | TV-L1 | AR-NLM | LRA-SVD | NCSR |
|---|---|---|---|---|---|---|---|---|---|---|---|---|
| PSNR | Cameraman | 38.21 | 35.20 | 36.84 | 37.75 | 38.12 | 38.55 | 38.07 | 25.84 | 37.98 | 38.19 | 38.28 |
| | Lena | 38.66 | 37.10 | 37.64 | 38.78 | 38.85 | 39.51 | 39.12 | 29.23 | 38.04 | 39.32 | 39.31 |
| | Barbara | 38.27 | 33.71 | 36.85 | 37.59 | 37.44 | 38.94 | 38.36 | 28.53 | 37.41 | 38.65 | 38.40 |
| | Boat | 37.21 | 36.64 | 35.64 | 37.24 | 37.46 | 37.93 | 37.37 | 26.30 | 37.04 | 37.58 | 37.69 |
| | Couple | 37.44 | 36.38 | 36.00 | 37.29 | 37.41 | 37.88 | 37.26 | 27.36 | 36.79 | 37.54 | 37.69 |
| | Fingerprint | 36.47 | 34.72 | 35.82 | 35.28 | 35.03 | 35.59 | 35.50 | 19.96 | 26.97 | 35.44 | 35.40 |
| | Hill | 37.10 | 36.21 | 35.42 | 37.08 | 37.20 | 37.51 | 37.17 | 29.35 | 36.02 | 37.26 | 37.39 |
| | House | 39.80 | 37.76 | 37.12 | 39.42 | 38.91 | 39.99 | 39.51 | 32.43 | 39.00 | 39.94 | 39.83 |
| | Man | 37.77 | 36.21 | 36.00 | 36.95 | 37.20 | 37.61 | 37.14 | 25.95 | 36.75 | 37.28 | 37.39 |
| | Montage | 41.07 | 38.51 | 39.31 | 39.91 | 40.34 | 41.46 | 40.54 | 24.29 | 40.34 | 40.70 | 40.99 |
| | **Average** | **38.20** | **36.25** | **36.66** | **37.73** | **37.80** | **38.50** | **38.01** | **26.92** | **36.63** | **38.19** | **38.24** |
| Metric | Image | BM3D | DnCNN | KLLD | PLOW | EPLL | WNNM | LPG-PCA | TV-L1 | AR-NLM | LRA-SVD | NCSR |
| SSIM | Cameraman | 0.9612 | 0.93862 | 0.6067 | 0.9561 | 0.9616 | 0.9623 | 0.9601 | 0.8186 | 0.9586 | 0.9613 | 0.9605 |
| | Lena | 0.9711 | 0.9503 | 0.6680 | 0.9671 | 0.9690 | 0.9722 | 0.9705 | 0.8884 | 0.9625 | 0.9719 | 0.9715 |
| | Barbara | 0.9723 | 0.9129 | 0.8040 | 0.9666 | 0.9662 | 0.9751 | 0.9725 | 0.8472 | 0.9657 | 0.9740 | 0.9724 |
| | Boat | 0.9658 | 0.9570 | 0.7105 | 0.9617 | 0.9656 | 0.9678 | 0.9641 | 0.7947 | 0.9609 | 0.9659 | 0.9656 |
| | Couple | 0.9675 | 0.9546 | 0.7854 | 0.9642 | 0.9661 | 0.9686 | 0.9649 | 0.7925 | 0.9554 | 0.9673 | 0.9673 |
| | Fingerprint | 0.9897 | 0.9890 | 0.9837 | 0.9900 | 0.9895 | 0.9904 | 0.9902 | 0.6028 | 0.9079 | 0.9902 | 0.9901 |
| | Hill | 0.9577 | 0.9470 | 0.7820 | 0.9552 | 0.9582 | 0.9589 | 0.9569 | 0.8237 | 0.9363 | 0.9584 | 0.9579 |
| | House | 0.9554 | 0.9336 | 0.4893 | 0.9524 | 0.9492 | 0.9576 | 0.9552 | 0.8623 | 0.9490 | 0.9599 | 0.9572 |
| | Man | 0.9619 | 0.9490 | 0.7756 | 0.9578 | 0.9619 | 0.9640 | 0.9606 | 0.8363 | 0.9557 | 0.9622 | 0.9619 |
| | Montage | 0.9817 | 0.9660 | 0.6885 | 0.9739 | 0.9795 | 0.9825 | 0.9804 | 0.9152 | 0.9785 | 0.9807 | 0.9821 |
| | **Average** | **0.9684** | **0.9498** | **0.7294** | **0.9645** | **0.9667** | **0.9699** | **0.9675** | **0.8182** | **0.9531** | **0.9692** | **0.9686** |

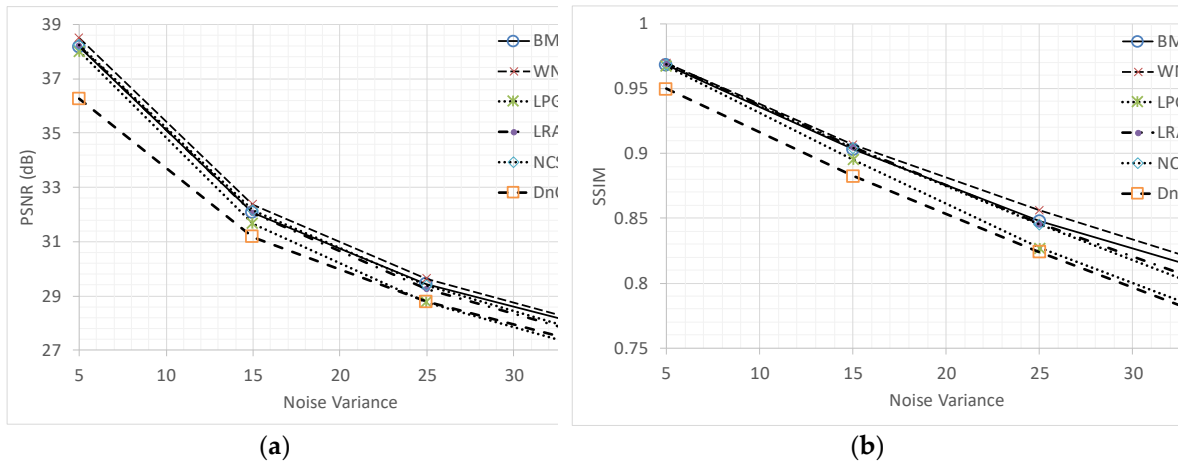

**Figure 1.** Comparison of image denoising performance at different noise variances (a) based on PSNR (b) based on SSIM.

## 2.2. Theoretical foundations of the denoising algorithms

Let Y denote an observed noisy image defined as $Y = Z + E$, where Z is the noise free image and E represents the noise. The goal of image denoising is to approximate Z, given Y. Often in literature E is modelled as Additive White Gaussian Noise (AWGN).

Image denoising is a classic ill-posed problem, where an ill-posed problem is defined as a problem: that may not have solution due to the presence of noise, for which may not exist a unique solution, and for which the solution may not be stable under small perturbations to the observations.

In this section we will provide a tutorial overview of the state-of-the-art algorithms for image denoising. We start by revisiting the popular Non-Local Means (NLM) algorithm [4] and follow it up with major algorithms that followed a similar philosophy.

The NLM algorithm tries to de-noise an image by replacing the pixels values with a weighted average of the values of its neighbouring pixels. In this algorithm, the weighting of each neighbouring pixel is derived by considering the structural similarity of each neighbour. To calculate the structural similarity, a patch if the image centred at each pixel is considered. The main supporting assumption here is that every patch in an image has structurally similar patches elsewhere within the same image [4].

Given a noisy image $Y = \{y_i | i \in I\}$, the denoised value $\hat{y}_i$ of pixel $i$, is computed as the weighted average of all the pixels in the image,

$$\hat{y}_i = \sum_{j \in I} w_{i,j} \cdot y_j, \qquad (1)$$

Where $w_{i,j}$ is the structural similarity between the current pixel $i$, and neighbouring pixel j,

$$w_{i,j} = \frac{1}{Z(i)} e^{-\frac{\|N_i - N_j\|_{2,a}^2}{h^2}}, \text{ and } Z(i) = \sum_j e^{-\frac{\|N_i - N_j\|_{2,a}^2}{h^2}} \qquad (2)$$

where $N_k$ denotes the intensity gray levels vector of the square neighbourhood of fixed size centered at k. a is the standard deviation of the Gaussian kernel and h is a smoothing factor.

Since the introduction of NLM filter as a mode of denoising, a multitude of algorithms have been proposed in literature. Most of the algorithms exploit the philosophy that there exists a structural redundancy in images. However, most of the algorithms attempt to denoise an image patch at once rather than to process individual pixels as proposed in NLM.

At a higher level, the algorithms presented in literature is constituent of 4 major components. To denoise an image patch (collection of pixels), the first step of all the algorithms is to identify structurally similar pixels or image patches. The second step is to represent these patches as a linear combination of a set of basis functions. Almost all the algorithms employ regularization as a technique to isolate image structures from noise details. Finally, most of the algorithms use a boosting technique to improve an initial denoised estimate. In addition to the above four components, some of the papers also describe a method to estimate the noise level at the beginning. In the following sections, we describe each of these techniques while highlighting the major differences between competing algorithms.

*2.3. Major components of patch similarity based denoising algorithms*

1) Identification of similar image patches (or pixels)

Let's denote the image patch $x^j$ ($j = 1, 2, \ldots m$) of size $\sqrt{n} \times \sqrt{n}$ centred at pixel $y_i$, is vectorised as $X^j$, where $X^j$, is represented as a n-component vector.

$$X^j = [x_1^j, x_2^j, x_3^j, \ldots \ldots x_n^j]^T \qquad (3)$$

The $x_i^j$, where $i \in 1..n$, in equation (1) denotes the pixels of patch $x^j$. The patch $x^j$ is compared against all the overlapping patches in the neighbourhood of $y_i$. The similarity between $x^j$ and a patch $x^k$ in the neighbourhood is calculated as a distance measure. The simplest of the distances measure is as follows:

$$S(x^j, x^k) = \|x^j - x^k\|_2^2 \qquad (4)$$

In certain algorithms all the patches in the neighbourhood is considered for the patch based denoising, and in some other algorithms such as BM3D, only a subset of the patches in the neighbourhood are selected. There could be different approaches to select a subset of patches from the neighbourhood. The first approach is based on a selected threshold, where all the patches for which the distance measure is less than the selected threshold \theta selected. The second approach is to select the closest patches based on the distance measure.

All the image patches deemed similar to $x^j$ are concatenated as a matrix $X \epsilon \mathcal{R}^{n \times m}$ as follows,

$$X = \begin{bmatrix} x_1^1 & x_1^2 & \cdots & x_1^m \\ x_2^1 & x_2^2 & \cdots & x_2^m \\ \vdots & \vdots & \vdots & \vdots \\ x_n^1 & x_n^2 & \cdots & x_n^m \end{bmatrix}, \tag{5}$$

Where $m$ is the number of patches similar to the patch centred at $y_i$.

2) Representation in a linear basis

The objective of this step is to represent an image patch or a set of image patches, denoted below as X, as a linear combination of a representation basis as follows,

$$X = \sum_{i=1}^{N} a_i \varphi_i, \tag{6}$$

Where $\varphi_i$ ($i = 1, 2, \ldots N$) are the basis functions, and $a_i$ ($i = 1, 2, \ldots N$) are the representation coefficients. In practice $\varphi_i$ could be selected as fixed basis such as wavelets or Discrete Cosine Transform (DCT) or can be adaptive to the content. In the popular BM3D algorithm DCT is selected along with Haar wavelets as the basis to represent the noisy image patches.

When the representation basis is adapted to the content at hand, it is commonly known as a dictionary. Two of the most popular dictionary learning methods are the Principle Component Analysis (PCA) and Singular Value Decomposition (SVD).

To calculate the PCA, the matrix X is centralized by subtracting the mean of each vector component,

$$\bar{X} = X - E(X), \tag{7}$$

Where, $E(x) = \{\mu_i, i = 1 \ldots n\}$, and $\mu_i = 1/m \sum_{j=1}^{m} x_i^j$.

The PCA is then obtained by taking the eigenvalue decomposition of the covariance matrix (which is a symmetric matrix) of the centralized dataset as follows,

$$\Omega = \frac{1}{m} \overline{XX}^T = \Phi \Lambda \Phi^T, \tag{8}$$

where $\Phi$ is the $n \times n$ orthonormal eigenvector matrix and $\Lambda$ is the diagonal eigenvalue matrix. Now $\Phi$ can be used as an orthonormal transformation matrix to decorrelate $\bar{X}$, as $\bar{Y} = \Phi^T \bar{X}$, where $\Lambda = \frac{1}{m} \overline{YY}^T$. The original dataset $\bar{X}$ is now transformed into $\bar{Y}$, where the signal energy is concentrated on a small subset of $\bar{Y}$, while the energy of the noise is evenly spread over the whole data set. The LPG – PCA algorithm [17] utilizes the PCA as its representation basis.

The other most popular matrix decomposition is known as the Singular value decomposition (SVD). Here instead of decomposing the covariance matrix, the centralized data set is decomposed as follows,

$$\bar{X} = U\Sigma V^T = \sum_{i=1}^{m} \sigma_i u_i v_i^T \tag{9}$$

Where, $= (u_1, \ldots, u_m) \epsilon \mathcal{R}^{n \times m}$, and $V = (v_1, \ldots, v_m) \epsilon \mathcal{R}^{m \times m}$, are orthonormal matrices, and $\sigma_i$ of $\Sigma$ are called the singular values of $\bar{X}$. It can be shown that PCA and SVD are related and does essentially the same sort of decorrelation of the dataset [40]. SVD is the representation basis used in LRA-SVD [7] and SAIST [26] algorithms.

An interesting development in the denoising algorithms of the past decade involved sparse linear representations of image patches. In such algorithms image patches are described by sparse linear combinations of prototype-signal atoms from an over complete dictionary. Learning such content adaptive sparse representation basis is a challenge by itself. The goal of such an approach is to learn a dictionary D of k atoms in R[Nxk], for an image of size N, with N overlapping patches of size m, with representation vectors $\alpha_i$, and a target accuracy $\epsilon$.

**Table 3:** Comparison of building blocks of image denoising algorithms

| Algorithm | Identification of similar patches | Representation Basis | Regularization | Boosting |
|---|---|---|---|---|
| NLM | All overlapping patches in the neighborhood considered, but weighted by the negative exponential of the Euclidian distance between the patches | Pixel domain | None | [8] None |
| BM3D | Coarse pre-filtering of the patches and the Euclidean distance between the transform coefficients is considered. | Discrete Cosine Transform | Wiener Filter, with an initial estimate of the image obtained through hard thresholding of transformed patches | None |
| K-SVD | | Globally trained dictionary | Thresholding of coefficients depending on the noise level | None |
| PLOW | Geometric clustering, using LARK features, followed by photometric similarity | Pixel domain | Wiener filter, with an initial estimate obtained with a lower noise variance estimate | None |
| EPLL | MAP estimation | Gaussian Mixture Model based prior | Wiener filter | Iterative filtering |
| LPG-PCA | Euclidean distance between the patches computed and any patch less than a threshold is selected | PCA | Shrinking the eigenvalues proportional to the estimated noise variance | Two stage filtering. |
| AR-NLM | Similar to NLM, with a minor variation on the exponential decay function | Pixel Domain | Competition between non-local means and total variation regularization | None |
| LRA-SVD | Euclidean distance between the patches computed to find the closest 85 patches | SVD calculated on the group of patches | Low rank approximation by comparing the singular values to noise variance | Twicing: add a proportion of filtered noise back to the denoised image and repeat |
| NCSR | Cluster the image patches in to 70 clusters with k-means clustering | PCA sub-dictionary learnt for each cluster | Grouped sparsity regularization by modelling sparse coding noise | Iterative thresholding [35] |
| SAIST | Not clear from the paper but assume same as NCSR | SVD calculated on the group of patches | Soft thresholding of the singular values [36], with threshold derived similar to [8] | Iterative filtering, with a proportion of noise added back |
| LSSC | Not stated, but grouping is done after one round of denoising. | A dictionary learnt offline from $2 \times 10^7$ image patches | Grouped sparsity regularization: Forcing a common sparsity pattern among similar patches | None |
| WNNM | Not stated, assumed similar to NLM | SVD calculated on the group of patches | Shrinking singular values (SV) proportional to the size of the SV | Iterative filtering, with a proportion of noise added back |

$$\min_{D \in \mathbb{C}, A} \sum_{i=1}^{N} \|\alpha_i\|_p \ s.t. \ \|z_i - D\alpha_i\|_2^2 < \epsilon \qquad (10)$$

Where $\mathbb{C}$ is the set of matrices in $\mathcal{R}^{N \times k}$ with unit $l_2$-norm columns, A = $[\alpha_1, \alpha_2, \ldots \alpha_N]$ is a matrix in $\mathcal{R}^{k \times m}$, $z_i$ is the i[th] patch of noisy image $Y$.

The K-SVD algorithm is a popular method of learning over complete dictionaries for tasks that involve sparse representations [41]. Due to the high computational complexity involved, it is common to learn over complete dictionaries offline using training images. The EPLL algorithm learns such a dictionary (referred to as patch priors) with a Gaussian mixture model (GMM) of 200 mixture components from a set of 2 million image patches sampled from an image data base.

3) Regularization

As introduced in the beginning denoising is an ill-posed problem. Regularization is the method to deal with all the problems associated with coming up with a solution to such ill-posed problems. Furthermore, regularization paves way to incorporate prior knowledge about the domain in to the solution [42]. As such, the regularization component of the algorithms is formulated to include a data fidelity term and a side constraint term that captures the prior knowledge of the behaviour of the solution.

The most commonly used method of regularization is a low rank approximation of the dataset in the basis representation. This method when used along with SVD representation is known as Truncated SVD regularization. Here, a low rank approximation of $\overline{X}$, $\overline{X}_{LRA}$ of rank r, is found by setting the $m - r$ smallest singular values of (**9**) to be equal to zero as follows,

$$\sigma_1 \geq \sigma_2 \geq \cdots \ldots \sigma_r = \sigma_{r+1} = \cdots \ldots \sigma_m = 0, \tag{11}$$

$$\overline{X}_{LRA} = \sum_{i=1}^{r} \sigma_i u_i v_i^T,$$

The assumption here is that the noise energy is spread equally among all the basis dimensions, whereas the signal energy (the useful component) is concentrated in few basis dimensions. The question lies in how to select an appropriate rank to represent $\overline{X}_{LRA}$. In [7], r is found by the following criteria,

$$\sum_{i=r}^{m} \sigma_i^2 > \tau^2 \geq \sum_{i=r+1}^{m} \sigma_i^2, \text{where } \|\overline{X} - \overline{X}_{LRA}\|_F^2 = \tau^2 \tag{12}$$

Another very popular method of regularization is the Tikhonov regularization. The Tikhonov regularization is formulated to find a trade-off between the data fidelity and the energy in the solution as follows,

$$\overline{A}_{Tik} = \arg\min_{A} \|\overline{X} - A\Phi\|_2^2 + \mu^2 \|A\|_2^2, \tag{13}$$

Where, A = $[\alpha_1, \alpha_2, \ldots \alpha_N]$ and $\Phi$ is the dictionary with basis vectors $\varphi_i$ ($i = 1, 2, \ldots N$). Under certain choices of $\mu$, Tikhonov regularization turns out to be essentially the same as the popular Wiener filter [42, p. 6]. For example, when the representation basis $\Phi$ utilised is 2D-DCT of the corresponding image patch (or patches), Wiener filter is the element-wise weighting of the DCT coefficients y_DCT, as follows,

$$y_{Wiener} = \left(\frac{S_y}{S_y + S_\eta}\right) \cdot y_{DCT}, \tag{14}$$

where $S_y$ is the power spectral density of the element corresponding to y_DCT and $S_\eta$ is the noise variance. The power spectral density is the 2D-DCT of the corresponding covariance matrix. Wiener filter is the preferred form of regularization in several algorithms including BM3D [15], PLOW [14] and LPG-PCA [2].

When used to reduce high frequency noise in the image, the quadratic regularization term in (**11**) has a tendency to suppress useful high frequency energy in the image too. Therefore, researchers have often used a non-quadratic regularization criteria for image denoising. In EPLL algorithm the regularization problem is formulated with criteria called the Expected Patch Log Likelihood (EPLL) as follows,

$$\arg\max_{Z} \frac{\lambda}{2} \|Z - Y\|^2 - EPLL_p(Z), \tag{15}$$

Where, $EPLL_p(Z) = \sum_i \log p(P_i Z)$, and $P_i$ is a matrix which extracts the patch centred at pixel y_i, and $\log p(P_i Z)$ is the log likelihood of the patch under the prior $p$.

Another class of regularization techniques used in image denoising work based on the assumption that image patches can be sparsely represented in certain domains. Thus, the regularization problem is formulated in search of different sparsity constraints. For example in [35], authors propose to include the l_0 norm of the representation vector as the regularization term,

$$A_{KSVD} = \arg\min_{A} \|\overline{X} - A\Phi\|_2^2 + \mu \|A\|_0, \text{ where, } \|A\|_0 = \sum_i^m \|\alpha_i\|_0 \tag{16}$$

In [18], authors go one step further to include another sparsity constraint. In this case authors argue that due to the correlations across the image patches that exist in images, the representation vectors $\alpha_i$ are also related to each other. Thus, sparse representation vectors $\alpha_i$ are centralized to an estimate $\beta_i$ and the regularization formulation is as follows,

$$A_{NCSR} = \arg\min_{A}\|\bar{X} - A\Phi\|_2^2 + \mu\|A\|_1 + \gamma\|A - B\|_p, \tag{17}$$

where $B=[\beta_1, \beta_2, \ldots \beta_N]$ is a good estimate of A, and p is selected as 1, or 2. $\beta_i$ is found using the weighted average of the sparse codes associated with non-local similar patches to patch at $y_i$.

Another approach to promote similar sparse representation vectors for patches in a group of patches, a group sparsity constraint is included in the regularization step. The approach known as simultaneous sparse coding is formulated as follows,

$$A_{SSC} = \arg\min_{A}\|\bar{X} - A\Phi\|_2^2 + \mu\|A\|_{p,q}', \text{ where } \|A\|_{p,q} = \sum_{i}^{m}\|\alpha_i\|_q^p. \tag{18}$$

Simultaneous sparse coding is implemented both in [36] and [26], where the adaptive basis (dictionary) is also learnt along with the representation vector,

$$(A_{SSC}, \Phi_{SSC}) = \arg\min_{A,\Phi}\|\bar{X} - A\Phi\|_2^2 + \mu\|A\|_{p,q}. \tag{19}$$

4) Aggregation

The purpose of regularization step was to find a solution to the ill-posed nature of the image denoising problem. The previous section illustrated how each overlapping patch was represented as a linear combination of some representation basis. Due to the overlapping nature of the patches, each pixel will have multiple estimates. The next step of the algorithms is to aggregate multiple estimates of each pixel.

5) Boosting of the initial denoised estimate

The result of the regularization step is dependent on the patch grouping. However, to identify similar patches the algorithm has to rely on the noisy input image. The noise in the original input image inadvertently affects the patch grouping step. Due to this reason, there is often, some noise that is present in the denoised image. Therefore, it is common for most algorithms to implement an iterative version of the denoising algorithm. The goal of this step is to further improve the performance of the denoising algorithm.

The repeated application of the denoising algorithm will pave way for better estimation of noise level in the image and improve on patch grouping. Furthermore, it is also common for the dictionaries to be updated on a recursive basis until convergence [18]. While it is possible to improve the results through repeated denoising, a more principled approach is to deal with the noise residual, which is the difference between the original noisy image and the denoised image [29].

One of the popular techniques of boosting is referred to as "twicing" tries to extract the signal leftovers in the noise residual and add it back to the denoised image. The signal leftovers in the noise residual are extracted by denoising the residual. The twicing technique can be summarized as follows,

$$\hat{Z}^{k+1} = \hat{Z}^k + f(Y - \hat{Z}^k), \tag{20}$$

Where $\hat{Z}^k$ is the denoised estimate at the k$^{th}$ iteration and Y is the noisy image, and $f(\cdot)$ is the denoising algorithm.

The other technique referred to as "back projection", works by adding the residual back in to the denoised image,

$$\hat{Y} = \hat{Z}^k + \delta(Y - \hat{Z}^k), \tag{21}$$

Where $\hat{Y}$ is the input to the second is stage of denoising, and $\delta\epsilon(0,1)$ is a relaxation parameter. An alternative proposal known as SOS (Strengthen, Operate and Subtract) boosting is proposed in [43], where the authors try to compensate for the "disagreement" between the local patch denoising and the global result obtained by the aggregation process with a global subtraction operation,

$$\hat{Z}^{k+1} = f(Y + \hat{Z}^k) - \hat{Z}^k, \tag{22}$$

*2.4. Theoretical comparison of the image denoising algorithms*

In the Table 3 we compare and contrast between different image denoising algorithm. For the comparison we consider 4 components: method for identification of similar patches, representation basis, regularization scheme and boosting methodology. Apart from these components, aggregation methodology is also an important component, but has very minor variations between the algorithms. Hence, we do not include that step for our comparison.

In Summary, the publications on image denoising that claim state-of-the-art performance is a concatenation of at least 4 distinct steps. As the above table illustrates, there is a plethora of techniques utilized at each step. Each of these steps is crucial for the ultimate performance of the denoising algorithm. While it is often common for the authors to present overall quality of the output, this leaves the reader wondering where the denoising gains (in terms of PSNR / or the quality metric) come from. For example, although the authors in [15], suggested that a data driven representation basis such as PCA/SVD would yield better results, we are yet to know which data driven basis is good. This is because this insight is often hidden within the overall picture. In our opinion, this is partly responsible for any algorithm to not be able to consistently and convincingly outperform BM3D algorithm. Therefore, focused research on individual steps, such as on Boosting in [43], might lead to break the plateau of performance of image denoising.

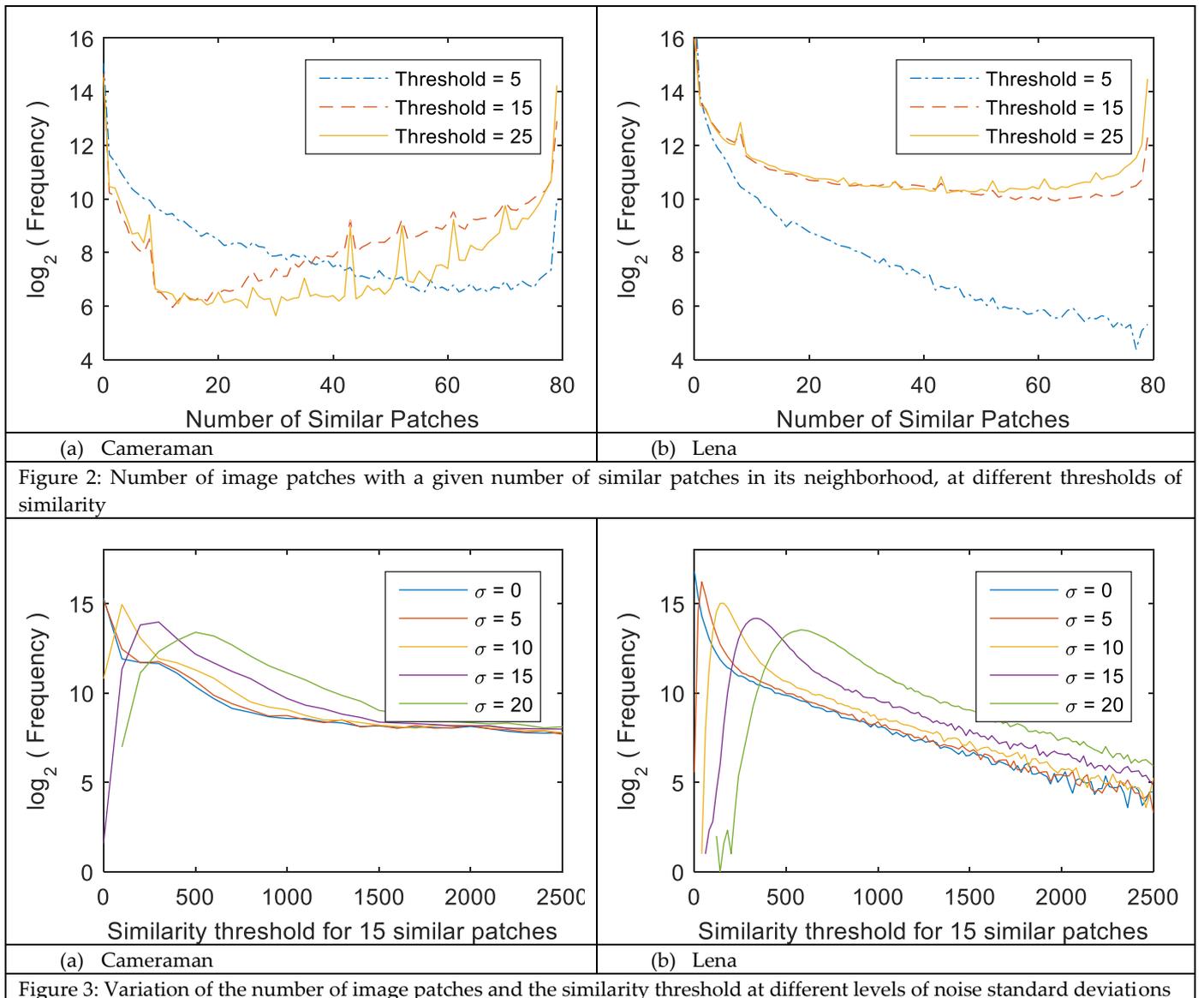

(a) Cameraman  (b) Lena

Figure 2: Number of image patches with a given number of similar patches in its neighborhood, at different thresholds of similarity

(a) Cameraman  (b) Lena

Figure 3: Variation of the number of image patches and the similarity threshold at different levels of noise standard deviations

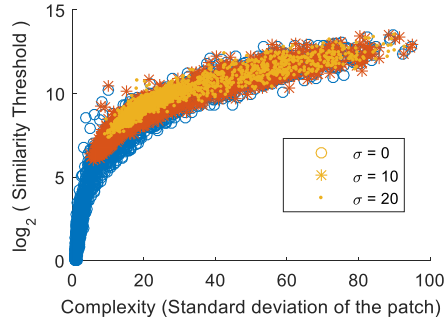

**Figure 4.** The similarity threshold against the complexity of a given patch at different noise levels.

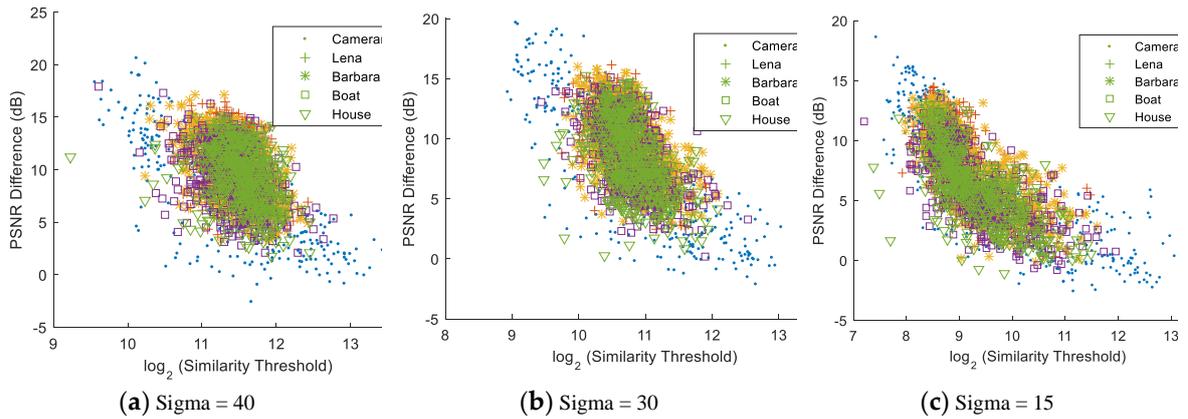

**Figure 5.** The PSNR difference (noisy-denoised) againt the similarlity threshold for a selection of image patches of different images.

## 3. Statistics of Structural Redundancy in natural images

A majority of image denoising algorithms that were developed in the recent past exploited the concept of non-local similarity. The concept of using structural similarity of pixels as weighting factor during image denoising was first proposed by Buades et al. [4]. The main proposition behind the idea of non-local similarity is that image patches demonstrate a spatial redundancy across the image. All of the state-of-the-art methods for image denoising use this property to find similar blocks of image patches. However, few questions arise as to the effectiveness of such a proposition. How likely is it to find similar patches to a given image patch? How does image noise affect the likelihood of finding similar patches? How far do we have to spatially traverse to find similar patches, or in other words how "non-local" do the patch similarities exist? How robust is this proposition under additive noise? Is there any relationship between contents of the patches to the statistics of non-local patch similarities? Do high amounts of patch similarities mean better denoising performance?

To the best of the authors' knowledge there is no statistical study related to the property of non-local similarity. In this section we try to answer some of the above questions through a detailed statistical study of the non-local similarities in natural images. The results of this investigation are organized under three subsections as follows.

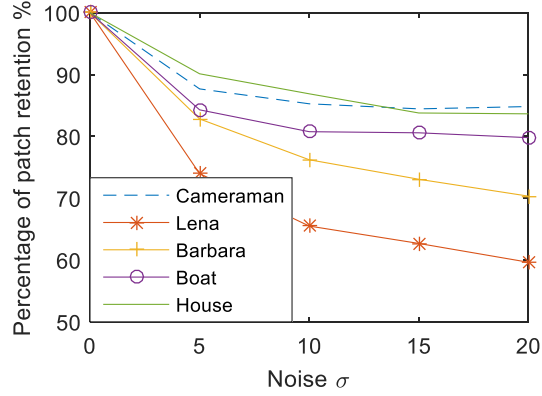

**Figure 6.** Percentage of patches retained (patches that are deemed similar to the patch that is to be denoised, when compared with no added noise) at different noise standard deviations.

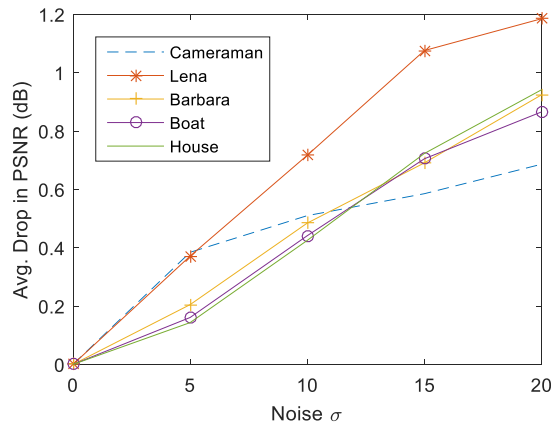

**Figure 7.** Average Drop in PSNR due to the difference in patch selection caused by noise.

*3.1. Global statistics of patch similarity*

Number To denoise an image patch, non-local similarity based denoising algorithms rely on identifying a number of similar patches in the neighbourhood. However, these algorithms suffer from the so called "rare patch effect" [39], where for some patches algorithm would not be able to find similar patches. In this section we analyse the global statistics related to patch similarity, to answer the question about "how likely is it to find similar patches to a given image patch?". This question, leads to another, "how related is this non-local similarity to the denoising performance?". To investigate towards an answer to above questions, we perform a couple of experiments. This section

In the first experiment, we count the number of similar patches for each patch centred at a given pixel in an image. This counting is performed on an image that has been denoised (without any noise added). We utilize a similarity measure as given in (1). A patch in the neighborhood is deemed similar if the similarity distance value is less than a given threshold. A patch of size 5x5 is considered in a search region of 9x9 pixels. The Figure 2 indicate the number of pixels (frequency: presented as the logarithm) with a given number of similar patches. As illustrated in Figures 2, at a very small threshold value, the frequency of similar patches decreases with the number of similar patches. However, as the threshold is increased, this observation is not held as more and more patches are deemed similar.

In the second experiment, we try to investigate the same relationship, when images are added with AWGN. In this case, we cannot do a like-to-like comparison between different noise levels, as the thresholds for similarity increases with the noise level. Therefore, to help with this scenario, we define a new measure of similarity called the similarity threshold. Similarity threshold is calculated by considering the 15 most similar patches based on (1), and taking highest similarity distance value among those patches. Thus, the similarity threshold value provides a measure of non-local similarity for a given pixel. The distribution of the similarity threshold of all the pixels in a given image is illustrated in Figure 3. As the noise level increases the distribution changes from an exponential distribution to a chi-squared distribution. The result of this is that higher the noise level, the number of pixels with non-local similarity tends to decrease. Does this mean, non-local similarity as a natural statistic becomes less effective as the noise increases? If so, at what rate does this effectiveness decline?

To investigate the above questions, we perform another experiment. In this experiment, 500 patches are chosen at random from 5 images added with AWGN with a noise standard deviation of 15. For each of the patches, all the patches from the neighbourhood are collected, and a patch based denoising is applied on the collection. The Figure 4 illustrates the similarity threshold against the complexity of the patches in this experiment. As the noise standard deviation increases, the similarity threshold tends to be concentrated at higher values. For denoising each group of patches, we utilize a similar approach to LRA-SVD [without the aggregation/iterative steps]. The PSNR of the resultant patches are plotted against the similarity threshold for the patch. The results of this experiment are presented in Figure 5. As the similarity threshold of a patch increases, the denoising performance measured as the PSNR decreases too. This illustrates that beyond a certain level of similarity threshold of a patch, patch based denoising is less effective.

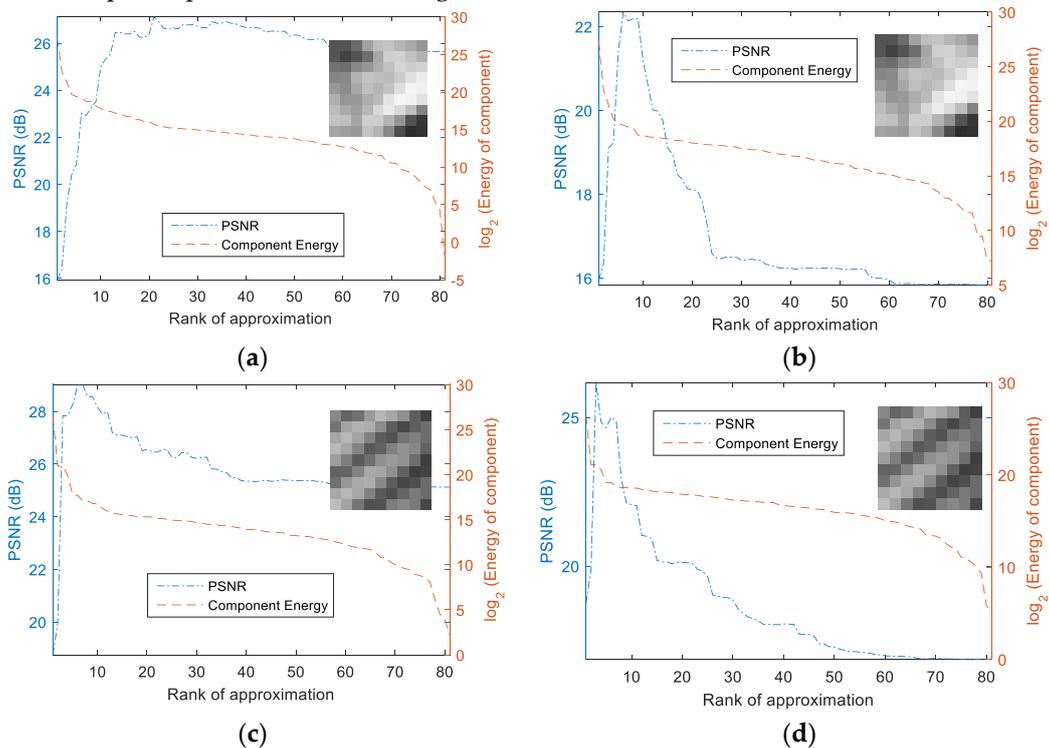

**Figure 8.** sparsity variation (a)/(c) Sigma 15, (b)/(d) Sigma 40.

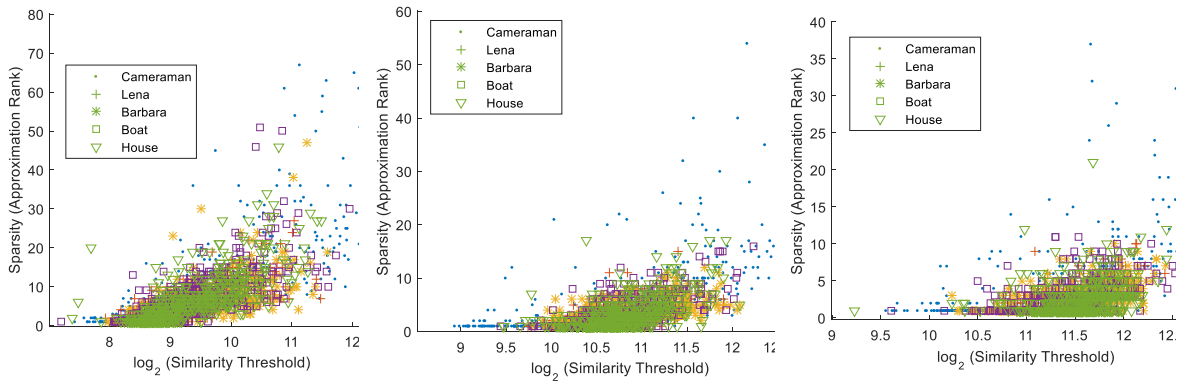

(**a**) (**b**) (**c**)

**Figure 9.** The variation of sparsity with similarity threshold at different levels of Sigma (a) 15, (b) 30, (c) 40.

*3.2. Robustness of patch similarity statistics under noise*

It is clear that in a noise free image there exists patch similarities in the local neighbourhood. However, when perturbed by noise, the algorithm inevitably finds patches that are incorrectly classed as being similar to the patch that is to be denoised. This is known in the literature as the "patch jittering effect" [39]. In this experiment we investigate the robustness of the patch similarity statistics under noise for a given patch similarity measure.

Similarity stats vs noise: In this experiment we select 100 patches along with its neighbourhood patches, from five image at random. Next, the 15 most similar patches are identified from the neighbourhood and the identity of those patches are stored. In the next step, the images are added with AWGN of varying noise levels and at each noise level, the 15 most similar patches are identified. In the final step, then identities of the similar patches at different noise levels, is compared with the identities of the similar patches when there was no noise added. Thus, we look at the percentage of retention of similar patches as compared to the when no AWGN is added. The results are illustrated in Figure 6.

So how does patch jittering affect the denoising performance? For this, we perform patch based denoising on the selected patches with 15 most similar patches identified, and compare it with denoising with the patches in the neighbourhood at similar locations as when there was no noise added. The PSNR difference is illustrated in Figure 7. According to the Figure 7, it is quite clear, that patch jittering has a negative effect on patch based denoising. Higher the jittering effect (i.e. lower the patch retention percentage), higher the drop in PSNR due to jittering. This effect gets worse as the noise variance increases, and hence curtailing the effectiveness of patch based denoising at higher noise levels.

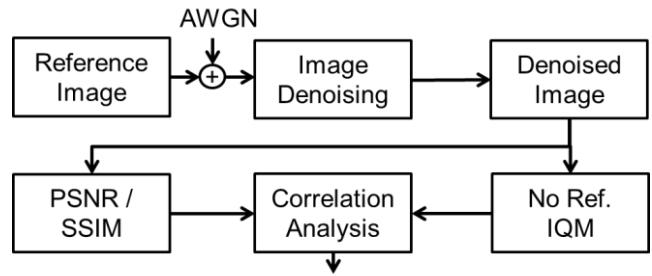

**Figure 10.** Experimental procedure to evaluate various NR-IQMs.

**Table 4.** Summary of correlation between various NR-image quality metrics and denoised image quality measured as SSIM and PSNR.

| Metric | SSIM | | |
|---|---|---|---|
| | CC | SSE | RMSE |
| BIQA | **0.6376** | 5.6093 | 0.1266 |
| BRISQUE | 0.3756 | 14.9445 | 0.2066 |
| NIQE | 0.2202 | 14.3026 | 0.2021 |
| IL-NIQE | 0.1894 | 13.3824 | 0.1955 |
| OG-IQA | 0.4224 | 7.7056 | 0.14838 |
| SHARPNESS | 0.5065 | 6.4188 | 0.1354 |
| Metric | PSNR | | |
| | CC | SSE | RMSE |
| BIQA | 0.2748 | 14492 | 6.4349 |
| BRISQUE | 0.0782 | 5192 | 3.8516 |
| NIQE | 0.076853 | 5908 | 4.1088 |
| IL-NIQE | 0.1939 | 3479 | 3.1528 |
| OG-IQA | 0.33104 | 4139 | 3.4392 |
| SHARPNESS | 0.1938 | 3805 | 3.2976 |

*3.3. Sparsity Analysis*

A common paradigm embraced by patch based denoising algorithms is to model a group of image patches as a sparse linear combination of prototype-signal atoms from an over complete dictionary. This sparsity assumption can be ascribed for the success of many imaging applications including image denoising. However, the question remains as to the level of sparsity that is observable in natural images. In particular, what is the level of sparsity observed in a group of patches that are deemed similar based on a given distance measure? Without this insight, hard thresholding operations become more commonplace in related imaging algorithms. To answer this question, we perform couple of experiments, which are described in the following text.

In the first experiment, we utilize 5 images that are already denoised (referred to as original), and corrupt them with AWGN. For each image, 500 random patches are extracted along with the overlapping patches from the neighbourhood. The experimental conditions: Noise standard deviation 15, patch size of 9x9 pixels, and a search window of 9x9. Thus, yielding 500 groups of patches for each image considered. For each neighbourhood patch within a group, the similarity is measured as given in (1). For each group of patches a dictionary is created through SVD as in [7]. The denoising of a group of patches is performed through the low rank approximation. For each group of patches, the rank of approximation is gradually increased, and at each increment, the PSNR of the patch is measured. The Figures 8 (a-e) illustrate the typical variation of PSNR of a denoised patch with the rank of approximation. Also, on the Figure 8, we illustrate the corresponding energy contained in each atom of the dictionary. The energy variation in the atoms corresponds to the similarity profile of the patches within the group. If most patches are similar the energy is contained within few atoms, and vice versa.

For the purpose of the discussions to follow, the sparsity observed in a group of patches is considered as the rank at which the PSNR of the denoised patch is the highest.

The Figures 9 (a,b,c) show the variation of the sparsity with the similarity threshold of the group, at different noise levels. Comparing Figures 9 a-c, when the noise level is low the sparsity observable in the group is low, which means that approximation rank is relatively higher. Furthermore, in general, as the similarity threshold increases, the approximation rank illustrates an increase as well.

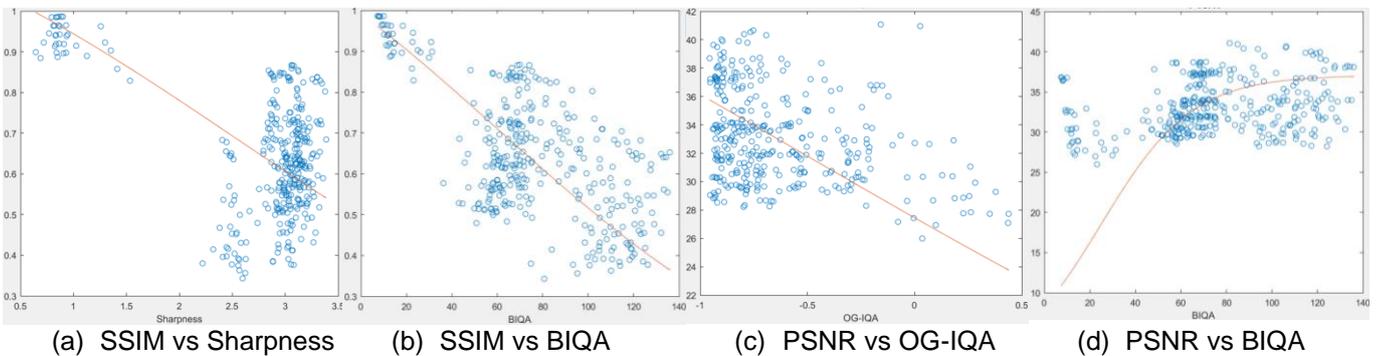

(a) SSIM vs Sharpness    (b) SSIM vs BIQA    (c) PSNR vs OG-IQA    (d) PSNR vs BIQA

Figure 11: Correlation of various No-Reference Image quality metrics with data fidelity and perceptual quality

## 4. Performance Evaluation of Image Denoising Algorithms and Future Directions

As hinted in the previous subsection, full reference image quality measurement techniques such as PSNR or SSIM are not suitable image denoising performance analysis. This is because we do not have the luxury of knowing the ground truth of the captured image (unless it is a synthetic image). While, subjective quality assessment is the best, the method used most often in the industry, for benchmarking purposes this is not very practical. Therefore, the development of bespoke NR image quality metrics for benchmarking image denoising performance is an open research need.

The objective of this section is to analyse the existing No-reference (NR) image quality metrics (IQM) to find a suitable metric to measure the quality of image denoising.

While we do not have the luxury of having the ground truth for captured images, for experimental purposes we use the standard set of images that were described in the previous experiments. We utilize the structural similarity index as a perceptual quality metric to measure the perceived quality of the denoised images, as compared to the reference image. While there are many NR IQMs, these are optimized to measure different kinds of distortions such as blur or blocking artefacts. Our objective is to select a NR-IQM suitable to measure the quality of denoised images, in particular denoised versions of images corrupted with AWGN. To select a suitable NR image quality metric, the correlation between different IQMs and SSIM and PSNR is analysed. The experimental procedure is illustrated in Figure 10.

For the purpose of this study we select 6 recently published no-reference (NR) image quality metrics (IQM). The NR-IQMs that are utilized in this study are as follows: Oriented Gradients Image Quality Assessment (OG-IQA)[44], the Blind/Reference-less Image Spatial Quality Evaluator (BRISQUE) index [45], Naturalness Image Quality Evaluator(NIQE) [46], A Feature-Enriched Completely Blind Image Quality Evaluator (IL-NIQE) [47], and global phase coherence based Sharpness Index (SHARPNESS) [48].

The PSNR will act as a data fidelity measure and SSIM will act as a perceptual quality measure. To create the test images for quality analysis, we use 10 different images, corrupted with AWGN at five different noise variance values (5,10,15,20,25) and utilize 7 different image denoising algorithms, resulting in 350 test images.

The correlation is measured after performing logistic regression of the metrics. The correlation plots for the best performing metrics in terms of data fidelity (PSNR vs NR-IQMs) and perceptual quality (SSIM vs NR-IQMs) are illustrated in Figure 11. The correlation results for all the measured NR-IQMs are summarized in Table 4.

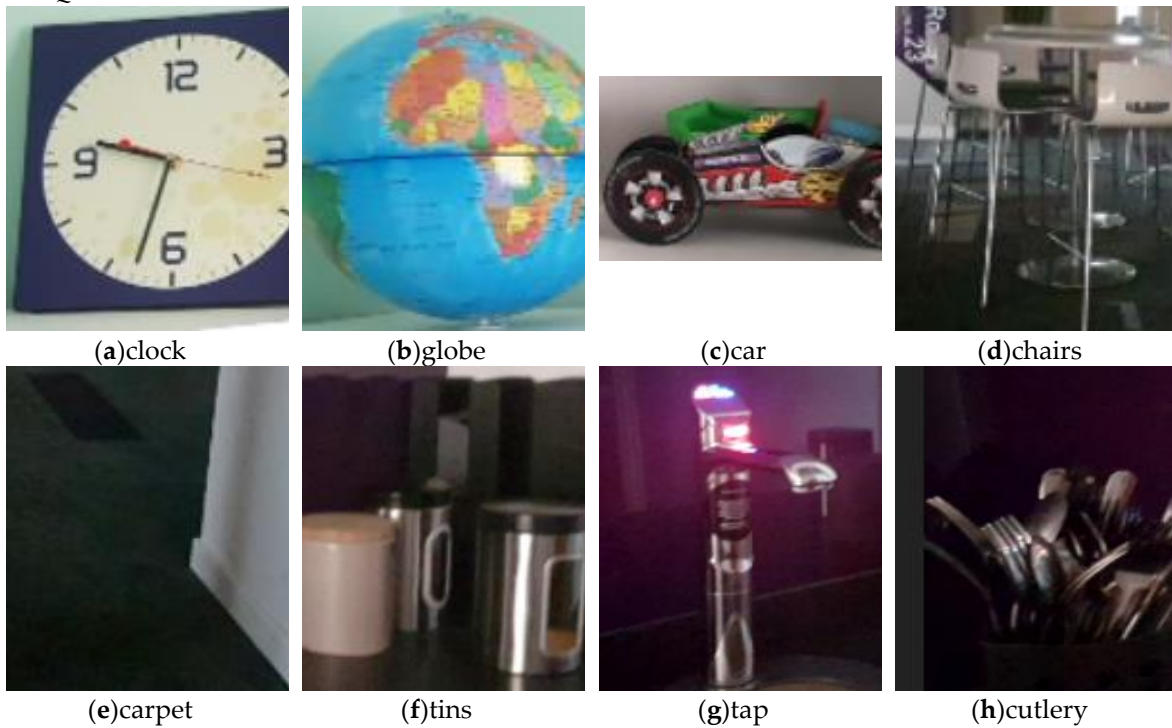

(**a**)clock  (**b**)globe  (**c**)car  (**d**)chairs
(**e**)carpet  (**f**)tins  (**g**)tap  (**h**)cutlery

**Figure 12.** Different Images Captured as RAW files that are used for quality comparison.

According to the results, BIQA metric performs the best in terms of predicting the perceptual quality of the denoised images. While the OG-IQA metric performs the best in terms of predicting data fidelity of denoised results, it has to be noted that the correlation is pretty poor between the PSNR and any of the tested NR metrics. The poor correlation with PSNR can be attributed to the fact that none of the NR metrics were targeted at measuring quality of denoised images.

Considering the results presented in Table 4, it is pretty clear that there is a requirement to develop novel no-reference quality metrics that are suitable for the purpose of denoising performance measurement. Of all the metrics considered only BIQA metric illustrates some correlation with the SSIM metric. Therefore, for the purpose of this study we will resort to the BIQA metric for measuring denoised image quality.

*4.1. Performance analysis with real raw sensor data*

The common methodology followed in image denoising literature is to add white Gaussian noise into an image, apply the denoising algorithm and compare between the denoised result and the original image. There is a flaw in this methodology, as the original image used in this scenario is already the result of a previous denoising operation. Thus, this image may contain residual noise that were not filtered by the denoising algorithm, and the method noise (new noise added into the image due to the denoising algorithm).

In this section we analyse the effects of this methodology. We use a set of raw images (with original acquisition noise) to apply the denoising algorithms and compare the performance among different denoising algorithms.

Image acquisition: To acquire the images used in this experiment we use the rear facing camera of a Samsung Galaxy S7 phone in the pro-mode. The raw images are produced in a ".DNG" format. The images are acquired in 4K mode at pixel resolution 2568x3768. A total of 8 images are captured and processed. A snapshot of the individual images is shown in Figure 12.

**Table 5**: No-Reference Performance Comparison of Denoising Algorithms on RAW images

| Image | Method | | | | | | |
|---|---|---|---|---|---|---|---|
| | BM3D | LRA/SVD | PLOW | NCSR | EPLL | WNNM | DnCNN |
| Clock | 83.11 | 72.32 | 75.97 | 67.09 | **122.87** | 76.65 | 81.01 |
| Globe | 146.94 | 147.60 | 143.56 | 158.36 | **176.43** | 171.72 | 138.24 |
| Car | 119.79 | 114.41 | 82.83 | **137.34** | 121.99 | 118.94 | 96.83 |
| Chairs | 74.95 | 77.02 | 72.23 | **88.45** | 58.63 | - | 74.11 |
| Carpet | 105.21 | 118.36 | 87.33 | 118.37 | 69.35 | **136.19** | 132.71 |
| Tins | 100.75 | 82.34 | 47.24 | 82.64 | 101.64 | 97.14 | **120.71** |
| Tap | 83.61 | 83.03 | 64.82 | 78.09 | 41.50 | 84.42 | **95.50** |
| Cutlery | 90.12 | 76.65 | 58.35 | 85.83 | 91.24 | - | **93.15** |

Raw image processing: To process the raw images captured above we implement the most basic features of an Image Signal Processor. The raw images are first denoised utilising the algorithm and then followed by Demosaicking and white balancing. Demosaicking is performed as a bi-cubic interpolation on a "GRGB" Bayer pattern. The outputs of the white balancing step are used as the input to the BIQA, NR image quality metric. The quality comparison results produced in Table 5, indicates that there is no clear winner as the ultimate image denoising algorithm. In most instances, a patch based image denoising method works better than DnCNN. The DnCNN algorithm performs well especially when the images are captured in dark environments. For most of the images considered, BM3D algorithm perform consistently well.

*4.2. Future Research Directions of Image Denoising*

The introductory experiments in section II illustrated that several patch based denoising algorithms perform at a similar level. WNNM algorithm marginally outperforms BM3D algorithm that was proposed more than a decade ago, when it came to comparisons of Gaussian noise removal. Interestingly, DnCNN algorithm did not perform satisfactorily in our experiments. In table 2, it was shown that there are very minor differences between this class of algorithms, which exploit the non-local similarity. When real noisy images are considered in Table 5, there was no clear winner. This leads to the question what else can be done to break this performance plateau.

Exploiting non-local similarity in an image was an interesting idea. Building dictionaries that represent the underlying image through non-local patches is the most critical step of all the algorithms considered in this paper. Authors use different sparsity priors to learn the dictionary atoms from non-local noisy patches. An alternative concept is to use not only non-local patches, but different non-noisy images to build these priors. This would address the rare-patch effect and the patch jittering effect explored in section III. Authors in [49], have explored this avenue at a comparatively basic level. This kind of activity is extremely difficult to implement though, as it would entail more than one image. However, deep learning algorithms for density estimation such as Generative Adversarial Networks (GANs)[50] is a promising approach to utilize for this scenario.

Another avenue to investigate is the development of no-reference image quality metrics and using such a metric to optimize the denoising process. Capturing raw images at different ISO levels to build representative ground truth images would be a good way to build a data set for this purpose [51]. Except for very few papers, comparing performance on real noisy images is still an aspect that is overlooked in published papers.

The patch based image denoising algorithms are made up of 5 distinct steps. They are grouping of similar patches from the neighbourhood, representation in a selected basis function, regularization/shrinkage of the coefficients, aggregation of the patches, and boosting of the initial estimate. While there are interesting variations among the techniques used in the above algorithmic steps, the published performance gains represent the overall denoising performance resulting from the totality of these steps. Hence, one cannot easily dissect where the performance gains came from. Therefore, it is suggested that it would be better if individual steps are compared and justified.

## 5. Conclusions

This paper described a study about a branch of image denoising algorithms that has achieved state-of-the-art performance, i.e. patch based image denoising algorithms that exploit non-local patch similarity. The paper started with a performance comparison based on the PSNR and SSIM metrics, which illustrated that there is very little performance gain achieved by the recently published algorithms over the popular BM3D algorithm, proposed in 2007. In search for reasons responsible for this performance plateau evident in the literature, this paper presented three avenues of investigation. A theoretical review of algorithms, scrutiny of the philosophy behind patch based denoising algorithms, and a critical evaluation of performance measurement methodology.

While there are interesting variations among the published techniques that use non-local similarity as a feature, these methods often have four steps that differentiate between the algorithms: selection of similar patches, representation learning, regularization and boosting. The published performance gains represent the overall denoising performance resulting from the totality of these steps, and it is not clear where the gains come from. Therefore, it might be timely to dissect these steps when research results are published.

The statistics of non-local patch similarity observed in natural images was also investigated in this paper. It is found that, the performance gains from patch based denoising decreases as the noise level increases. This reduction in performance gain is due to two main reasons: rare-patch effect (difficult to find similar patches from the neighbourhood), and patch jittering effect (identifying dissimilar patches as similar due to noise). Thus, the effectiveness of exploiting non-local similarity for denoising becomes less effective with increasing noise variance. It is often assumed in literature that group of patches can be represented in a sparse representation basis. Therefore, as part of this study we also analysed the level of sparsity observed in natural images under varying noise levels. To overcome the decreasing effectiveness of patch-based techniques under increasing noise, it is extremely necessary to come up with better patch similarity metrics that are robust under noise.

Furthermore, there is also a need for robust representation bases and adaptive regularization schemes that take in to account the variation in observed group sparsity under noise.

The finally the paper scrutinized the quality evaluation of denoising algorithms. Most papers evaluated the performance by adding AWGN noise to already denoised images and measure the PSNR or SSIM. In practice, we never have a reference to compare the performance against. Therefore, a no-reference image quality metric need to be utilized for this purpose. We compared the performance of existing no-reference image quality metrics for evaluation of image denoising. Utilizing the best among the no-reference quality metrics considered, we compared the performance of the image denoising algorithms on raw images captured by a CMOS camera. Therefore, developing such no-reference quality measurement techniques focused on image denoising is an important need. While such techniques would enable developers to benchmark their algorithms, it may also serve as an invaluable tool for adaptive regularization.

**Acknowledgments**: This work is supported by the Engineering and Physical Sciences Research Council (EPSRC) under the grant agreement EP/T000783/1

**Reference**

[1] C. Tomasi and R. Manduchi, "Bilateral Filtering for Gray and Color Images," *IEEE International Conference on Computer Vision, Washington DC, USA.*, pp. 839–846, 1998.

[2] G. Z. Yang, P. Burger, D. N. Firmin, and S. R. Underwood, "Structure adaptive anisotropic image filtering," *Image and Vision Computing*, vol. 14, no. 2, pp. 135–145, Mar. 1996.

[3] H. Takeda, S. Farsiu, and P. Milanfar, "Kernel Regression for Image Processing and Reconstruction," *IEEE Transactions on Image Processing*, vol. 16, no. 2, pp. 349–366, Feb. 2007.

[4] A. Buades, B. Coll, and J. M. Morel, "A non-local algorithm for image denoising," in *2005 IEEE Computer Society Conference on Computer Vision and Pattern Recognition (CVPR'05)*, 2005, vol. 2, pp. 60–65 vol. 2.

[5] E. P. Simoncelli and E. H. Adelson, "Noise removal via Bayesian wavelet coring," in *Proceedings of 3rd IEEE International Conference on Image Processing*, 1996, vol. 1, pp. 379–382 vol.1.

[6] J.-L. Starck, E. J. Candes, and D. L. Donoho, "The curvelet transform for image denoising," *IEEE Transactions on Image Processing*, vol. 11, no. 6, pp. 670–684, Jun. 2002.

[7] Q. Guo, C. Zhang, Y. Zhang, and H. Liu, "An Efficient SVD-Based Method for Image Denoising," *IEEE Transactions on Circuits and Systems for Video Technology*, vol. 26, no. 5, pp. 868–880, May 2016.

[8] S. G. Chang, B. Yu, and M. Vetterli, "Adaptive wavelet thresholding for image denoising and compression," *IEEE Transactions on Image Processing*, vol. 9, no. 9, pp. 1532–1546, Sep. 2000.

[9] A. Pizurica and W. Philips, "Estimating the probability of the presence of a signal of interest in multiresolution single- and multiband image denoising," *IEEE Transactions on Image Processing*, vol. 15, no. 3, pp. 654–665, Mar. 2006.

[10] L. Sendur and I. W. Selesnick, "Bivariate shrinkage functions for wavelet-based denoising exploiting interscale dependency," *IEEE Transactions on Signal Processing*, vol. 50, no. 11, pp. 2744–2756, Nov. 2002.

[11] T. Thaipanich, B. T. Oh, P. H. Wu, D. Xu, and C. C. J. Kuo, "Improved image denoising with adaptive nonlocal means (ANL-means) algorithm," *IEEE Transactions on Consumer Electronics*, vol. 56, no. 4, pp. 2623–2630, Nov. 2010.

[12] A. Wong, P. Fieguth, and D. Clausi, "A perceptually adaptive approach to image denoising using anisotropic non-local means," in *2008 15th IEEE International Conference on Image Processing*, 2008, pp. 537–540.

[13] D. V. D. Ville and M. Kocher, "Nonlocal Means With Dimensionality Reduction and SURE-Based Parameter Selection," *IEEE Transactions on Image Processing*, vol. 20, no. 9, pp. 2683–2690, Sep. 2011.

[14] P. Chatterjee and P. Milanfar, "Patch-Based Near-Optimal Image Denoising," *IEEE Transactions on Image Processing*, vol. 21, no. 4, pp. 1635–1649, Apr. 2012.

[15] K. Dabov, A. Foi, V. Katkovnik, and K. Egiazarian, "Image Denoising by Sparse 3-D Transform-Domain Collaborative Filtering," *IEEE Transactions on Image Processing*, vol. 16, no. 8, pp. 2080–2095, Aug. 2007.

[16] K. Dabov, A. Foi, V. Katkovnik, and K. Egiazarian, "BM3D Image Denoising with Shape-Adaptive Principal Component Analysis," presented at the SPARS'09 - Signal Processing with Adaptive Sparse Structured Representations, 2009.

[17] L. Zhang, W. Dong, D. Zhang, and G. Shi, "Two-stage image denoising by principal component analysis with local pixel grouping," *Pattern Recognition*, vol. 43, no. 4, pp. 1531–1549, Apr. 2010.

[18] W. Dong, L. Zhang, G. Shi, and X. Li, "Nonlocally Centralized Sparse Representation for Image Restoration," *IEEE Transactions on Image Processing*, vol. 22, no. 4, pp. 1620–1630, Apr. 2013.

[19] Y. Chen and T. Pock, "Trainable Nonlinear Reaction Diffusion: A Flexible Framework for Fast and Effective Image Restoration," *IEEE Trans. Pattern Anal. Mach. Intell.*, vol. 39, no. 6, pp. 1256–1272, Jun. 2017.

[20]  H. C. Burger, C. J. Schuler, and S. Harmeling, "Image denoising: Can plain neural networks compete with BM3D?," in *2012 IEEE Conference on Computer Vision and Pattern Recognition*, 2012, pp. 2392–2399.

[21]  K. Zhang, W. Zuo, Y. Chen, D. Meng, and L. Zhang, "Beyond a Gaussian Denoiser: Residual Learning of Deep CNN for Image Denoising," *IEEE Transactions on Image Processing*, vol. 26, no. 7, pp. 3142–3155, Jul. 2017.

[22]  C. Chen, Q. Chen, J. Xu, and V. Koltun, "Learning to See in the Dark," in *Computer Vision and Pattern Recognition, 2018. CVPR 2018. IEEE Conference on*, 2018.

[23]  I. Hong, Y. Hwang, and D. Kim, "Efficient deep learning of image denoising using patch complexity local divide and deep conquer," *Pattern Recognition*, vol. 96, p. 106945, Dec. 2019.

[24]  Z. Chen, Z. Zeng, H. Shen, X. Zheng, P. Dai, and P. Ouyang, "DN-GAN: Denoising generative adversarial networks for speckle noise reduction in optical coherence tomography images," *Biomedical Signal Processing and Control*, vol. 55, p. 101632, Jan. 2020.

[25]  Y. Zhong, L. Liu, D. Zhao, and H. Li, "A generative adversarial network for image denoising," *Multimed Tools Appl*, May 2019.

[26]  W. Dong, G. Shi, and X. Li, "Nonlocal Image Restoration With Bilateral Variance Estimation: A Low-Rank Approach," *IEEE Transactions on Image Processing*, vol. 22, no. 2, pp. 700–711, Feb. 2013.

[27]  A. Buades, B. Coll, and J. Morel, "A Review of Image Denoising Algorithms, with a New One," *Multiscale Model. Simul.*, vol. 4, no. 2, pp. 490–530, Jan. 2005.

[28]  J. I. de la R. Vargas, J. J. Villa, E. González, and J. Cortez, "A tour of nonlocal means techniques for image filtering," in *2016 International Conference on Electronics, Communications and Computers (CONIELECOMP)*, 2016, pp. 32–39.

[29]  P. Milanfar, "A Tour of Modern Image Filtering: New Insights and Methods, Both Practical and Theoretical," *IEEE Signal Processing Magazine*, vol. 30, no. 1, pp. 106–128, Jan. 2013.

[30]  L. Shao, R. Yan, X. Li, and Y. Liu, "From Heuristic Optimization to Dictionary Learning: A Review and Comprehensive Comparison of Image Denoising Algorithms," *IEEE Transactions on Cybernetics*, vol. 44, no. 7, pp. 1001–1013, Jul. 2014.

[31]  M. H. Alkinani and M. R. El-Sakka, "Patch-based models and algorithms for image denoising: a comparative review between patch-based images denoising methods for additive noise reduction," *EURASIP Journal on Image and Video Processing*, vol. 2017, p. 58, Aug. 2017.

[32]  B. Goyal, A. Dogra, S. Agrawal, B. S. Sohi, and A. Sharma, "Image denoising review: From classical to state-of-the-art approaches," *Information Fusion*, vol. 55, pp. 220–244, Mar. 2020.

[33]  V. Davidoiu, L. Hadjilucas, I. Teh, N. P. Smith, J. E. Schneider, and J. Lee, "Evaluation of noise removal algorithms for imaging and reconstruction of vascular networks using micro-CT," *Biomed. Phys. Eng. Express*, vol. 2, no. 4, p. 045015, 2016.

[34]  C. Tian, Y. Xu, L. Fei, and K. Yan, "Deep Learning for Image Denoising: A Survey," in *Genetic and Evolutionary Computing*, Singapore, 2019, pp. 563–572.

[35]  M. Elad and M. Aharon, "Image Denoising Via Sparse and Redundant Representations Over Learned Dictionaries," *IEEE Transactions on Image Processing*, vol. 15, no. 12, pp. 3736–3745, Dec. 2006.

[36]  J. Mairal, F. Bach, J. Ponce, G. Sapiro, and A. Zisserman, "Non-local sparse models for image restoration," in *Computer Vision, 2009 IEEE 12th International Conference on*, 2009, pp. 2272–2279.

[37]  P. Chatterjee and P. Milanfar, "Clustering-Based Denoising With Locally Learned Dictionaries," *IEEE Transactions on Image Processing*, vol. 18, no. 7, pp. 1438–1451, Jul. 2009.

[38]  X. Zhang, X. Feng, and W. Wang, "Two-Direction Nonlocal Model for Image Denoising," *IEEE Transactions on Image Processing*, vol. 22, no. 1, pp. 408–412, Jan. 2013.

[39]  C. Sutour, C. A. Deledalle, and J. F. Aujol, "Adaptive Regularization of the NL-Means: Application to Image and Video Denoising," *IEEE Transactions on Image Processing*, vol. 23, no. 8, pp. 3506–3521, Aug. 2014.

[40]  J. J. Gerbrands, "On the relationships between SVD, KLT and PCA," *Pattern Recognition*, vol. 14, no. 1, pp. 375–381, Jan. 1981.

[41]  M. Aharon, M. Elad, and A. Bruckstein, "$\rm K$-SVD: An Algorithm for Designing Overcomplete Dictionaries for Sparse Representation," *IEEE Transactions on Signal Processing*, vol. 54, no. 11, pp. 4311–4322, Nov. 2006.

[42]  W. C. Karl, "3.6 - Regularization in Image Restoration and Reconstruction," in *Handbook of Image and Video Processing (Second Edition)*, A. BOVIK, Ed. Burlington: Academic Press, 2005, pp. 183–V.

[43]  Y. Romano and M. Elad, "Boosting of Image Denoising Algorithms," *arXiv:1502.06220 [cs]*, Feb. 2015.

[44]  L. Liu, Y. Hua, Q. Zhao, H. Huang, and A. C. Bovik, "Blind image quality assessment by relative gradient statistics and adaboosting neural network," *Signal Processing: Image Communication*, vol. 40, pp. 1–15, Jan. 2016.

[45]  A. Mittal, A. K. Moorthy, and A. C. Bovik, "No-Reference Image Quality Assessment in the Spatial Domain," *IEEE Transactions on Image Processing*, vol. 21, no. 12, pp. 4695–4708, Dec. 2012.

[46] A. Mittal, R. Soundararajan, and A. C. Bovik, "Making a #x201C;Completely Blind #x201D; Image Quality Analyzer," *IEEE Signal Processing Letters*, vol. 20, no. 3, pp. 209–212, Mar. 2013.

[47] L. Zhang, L. Zhang, and A. C. Bovik, "A Feature-Enriched Completely Blind Image Quality Evaluator," *IEEE Transactions on Image Processing*, vol. 24, no. 8, pp. 2579–2591, Aug. 2015.

[48] G. Blanchet and L. Moisan, "An explicit sharpness index related to global phase coherence," in *2012 IEEE International Conference on Acoustics, Speech and Signal Processing (ICASSP)*, 2012, pp. 1065–1068.

[49] D.-V. Tran, S. Li-Thiao-Té, M. Luong, T. Le-Tien, and F. Dibos, "Patch-based image denoising: Probability distribution estimation vs. sparsity prior," in *2017 25th European Signal Processing Conference (EUSIPCO)*, 2017, pp. 1490–1494.

[50] I. J. Goodfellow *et al.*, "Generative Adversarial Networks," *arXiv:1406.2661 [cs, stat]*, Jun. 2014.

[51] T. Plotz and S. Roth, "Benchmarking Denoising Algorithms with Real Photographs," in *2017 IEEE Conference on Computer Vision and Pattern Recognition (CVPR)*, Honolulu, HI, 2017, pp. 2750–2759.